\documentclass[12pt]{article}
\usepackage[utf8]{inputenc}

\newcommand{\methodname}[1]{UVLN}

\usepackage{hyperref}
\usepackage{url}
\usepackage{booktabs}
\usepackage{graphicx}
\usepackage{times}
\usepackage{latexsym}
\usepackage{multirow}
\usepackage{multicol}

\usepackage{algorithm}
\usepackage{algorithmic}
\usepackage{colortbl}
\usepackage{amsmath}
\usepackage{amssymb}
\usepackage{soul, color, xcolor}
\usepackage{cite}

\title{Accessible Instruction-Following Agent}
\author{Kairui Zhou}
\date{Kent School, Connecticut, United States}

\begin{document}

\maketitle

\begin{abstract}
Humans can collaborate and complete tasks based on visual signals and instruction from the environment. Training such a robot is difficult especially due to the understanding of the instruction and the complicated environment.
Previous instruction-following agents are biased to English-centric corpus, making it unrealizable to be applied to users that use multiple languages or even low-resource languages.
Nevertheless, the instruction-following agents are pre-trained in a mode that assumes the user can observe the environment, which limits its accessibility.
In this work, we're trying to generalize the success of instruction-following agents to non-English languages with little corpus resources, and improve its intractability and accessibility.
We introduce UVLN (Universal Vision-Language Navigation), a novel machine-translation instructional augmented framework for cross-lingual vision-language navigation, with a novel composition of state-of-the-art large language model (GPT3) with the image caption model (BLIP).
We first collect a multilanguage vision-language navigation dataset via machine translation. Then we extend the standard VLN training objectives to a multilingual setting via a cross-lingual language encoder. The alignment between different languages is captured through a shared vision and action context via a cross-modal transformer, which encodes the inputs of language instruction, visual observation, and action decision sequences.
To improve the intractability, we connect our agent with the large language model that informs the situation and current state to the user and also explains the action decisions.
Experiments over Room Across Room Dataset prove the effectiveness of our approach. And the qualitative results show the promising intractability and accessibility of our instruction-following agent.

\end{abstract}

\section{Introduction}
The world we navigate through is multimodal and multilingual. A recent research topic, the Vision-Language Navigation task requires the agent to follow natural language instructions and navigate in houses, of which the main challenge is to handle multimodal input from the environment\cite{anderson2018vision, ku2020room, yan2019cross}.

Sequence-to-sequence architecture has dominated the implementation of the VLN task, where the instruction is encoded as a sequence of words, and the navigation trajectory is decoded as a sequence of actions\cite{wang2019reinforced, irshad2021hierarchical}. Sometimes, the attention mechanisms and beam search is applied to enhance the framework. While a number of previous methods have achieved success in language understanding, common to all existing work is that the agent learns to understand each instruction from single language, without leveraging multi-lingual knowledge.

However, the majority of current literature is biased towards English, making it imperfect for the existing paradigm of learning to understand the multi-lingual instruction. The problem is that, English-only VLN training is not good for generalization to following the instruction and single language model is difficult to adapt to multilingual environment. This is because ${1)}$ every instruction only partially characterizes the trajectory in single language. It can be difficult to adapt to other language, without grounding on the common language states. ${2)}$ The objects in multi language instructions may share various common relationships, and therefore it is natural to build an informative joint representation as common language knowledge for transfer learning in multilingual environment.

In general, there are three main possible challenges in multi-lingual VLN task, ${1)}$ Generalization to non-English or low-resource languages. ${2)}$  Interactively improve the performance of multilingual translation. ${3)}$ Bridge the semantic gap of different language instruction.

To address this natural difficulty of adapting to multilingual multimodal environment more effectively, we first collect multi-lingual dataset by applying machine translation (MT) module based on english-only dataset, and then propose to pre-train an encoder to align multiple language instructions and visual states for joint representations. The pre-trained model plays the role of providing generic image-text representations, and is applicable to most existing approaches to VLN.

\section{Related Works}
\subsection{Vision-Language Navigation}
Previous work ~\cite{hao2020learning} presents the first pre-training and fine-tuning paradigm for vision-and-language navigation (VLN) tasks. By training on a large amount of image-text-action triplets in a self-supervised learning manner, the pre-trained model provides generic representations of visual environments and language instructions.
In \cite{wang2019reinforced}, they propose a novel Reinforced Cross-Modal Matching (RCM) approach that enforces cross-modal grounding both locally and globally via reinforcement learning (RL) to address three critical challenges for this task: the cross-modal grounding, the ill-posed feedback, and the generalization problems.
In \cite{gan2020look}, they attempt to approach the problem of Audio-Visual Embodied Navigation, the task of planning the shortest path from a random starting location in a scene to the sound source in an indoor environment, given only raw egocentric visual and audio sensory data. In this work, we build on top of the recently proposed CLIP-ViL~\cite{shen2021much}, which achieves the state-of-the-art performance on the R2R benchmark~\cite{Anderson2018VisionandLanguageNI}. We further show that the CLIP-ViL model trained with monolingual data, namely the pivot method, cannot perform well in the multi-lingual 
settings, motivating the design of a specialized algorithm.

~\cite{fu2020counterfactual, Parvaneh2020CounterfactualVN} utilize the counterfactuals to augment either the trajectories or the visual observations.
~\cite{Liu2021VisionLanguageNW} proposes to mixup the environment.
~\cite{Majumdar2020ImprovingVN} leverages large-scale image-text pairs for the Web Data.
To leverage self-supervision, ~\cite{Zhu2020VisionLanguageNW} explores the design of self-supervised auxiliary tasks.
~\cite{Li2019RobustNW} explores the utilization of prior knowledge from the pre-trained models.
By training on a large amount of image-text-action triplets in a self-supervised learning manner, the pre-trained model provides generic representations of visual environments and language instructions.
Some studies ~\cite{hu2019looking, gan2020look} learn to utilize other available modalities.
~\cite{hao2020learning} presents pre-training and fine-tuning paradigm for vision-and-language navigation (VLN) tasks.
~\cite{wang2019reinforced} propose a novel Reinforced Cross-Modal Matching (RCM) approach that enforces cross-modal grounding both locally and globally via reinforcement learning (RL) to address three critical challenges for this task: the cross-modal grounding, the ill-posed feedback, and the generalization problems.

\subsection{Cross-modal Cross-lingual Learning}
Existing literature on Cross-modal Cross-lingual Learning mainly targets on two tasks: cross-modal retrieval~\cite{huang2020forward, burns2020learning, kim2020mule} and multimodal machine translation (MT)~\cite{calixto2017incorporating, calixto2017doubly, yao2020multimodal}. Using English as a pivot point, ~\cite{rajendran2015bridge, calixto2017sentence} pushes image features and semantic features of captions in different languages to the semantic feature of the English captions in the image captioning task. \cite{gella2017image} instead regards the image representation as the pivot, forcing representations of different languages to be close to the image representation. To scale to various downstream tasks \cite{zhou2021uc2} proposed the first machine translation-augmented framework for cross-lingual cross-modal representation learning by utilizing MT-enhanced translated data.
In this paper, we collect multi-lingual datasets by applying MT module based on English-only dataset, and aim to learn a cross-lingual cross-modal VLN agent that can effectively perform navigation under instructions from a diverse set of source languages. 

\subsubsection{Consistency Guidance}
~\cite{Xie2021PropagateYE, melaskyriazi2021pixmatch} learn dense feature representations via pixel-level visual consistency.
~\cite{Ouali2020SemiSupervisedSS, Abuduweili2021AdaptiveCR, Jeong2021FewshotOR} learn consistency regularization for semantic segmentation.
For example, ~\cite{Ouali2020SemiSupervisedSS} propose to maintain the context-aware consistency to make robust representations for varying environments.~\cite{Ren2021AdaptiveCP} proposes the Adaptive Consistency Prior for Image Denoising.
These studies expand the area of domain adaptation, transfer learning and few shot learning.
~\cite{Zhang_2021_CVPR} proposes to enforce the temporal consistency in low light video enhancement with static images.
~\cite{Ren2021AdaptiveCP} proposes an Adaptive Consistency Prior based Deep Network for Image Denoising.
Recent studies~\cite{Li20213DHA, Hyun2021SelfSupervisedVG, Yan2021SelfAlignedVD} show increasing interest of encouraging consistency across multiple modalities.
~\cite{Srinivas2020CURLCU} learns contrastive unsupervised representations for reinforcement learning.
~\cite{Hippocampus2020ASI} proposes to train the policy as a classifier via contrastive regularization for imitation learning.
In this study, we propose to leverage consistency into both imitation learning and reinforcement learning process of the VLN agent. 
Specifically, we utilize momentum contrast for enforcing visual consistency between query panoramic features as well as stable decision makings across similar observations.

\section{Problem Setup}

The vision-language navigation task requires an agent to find a route ${R}$ (a sequence of viewpoints) from the start viewpoint ${S}$ to the target viewpoint ${T}$ according to the given instruction ${I}$. The agent is put in a photo-realistic environment ${E}$. At each time step ${t}$, the agent's observation consists of a panoramic view and navigate viewpoints. The panoramic view ${o_t}$ is discretized into ${36}$ single views ${\{o_{t,i}\}_{i=1}^{36}}$. Each single view ${o_{t,i}}$ is an RGB image ${v_{t,i}}$ accompanied with its orientation ${(\theta_{t,i}, \phi_{t,i})}$, where ${\theta_{t,i}}$ and ${\phi_{t,i}}$ are the angles of heading and elevation, respectively. The navigable viewpoints ${\{l_{t,k}\}_{k=1}^{N_{t}}}$ are the ${N_t}$ reachable and visible locations from the current viewpoint. Each navigable viewpoint ${l_{t,k}}$ is represented by the orientation ${(\hat{\theta_{t,k}}, \hat{\phi_{t,k}})}$ from current viewpoint to the next viewpoints. The agents needs to select the moving action ${a_t}$ from the list of navigable viewpoints ${l_{t,k}}$ according to the given instruction ${I}$, history/current panoramic views ${\{o_{\delta}\}_{\delta=1}^{t}}$, and history actions ${\{a_\delta\}_{\delta=1}^{1:t-1}}$. In this paper, we consider the agent with full episode observability, all previous visual observations and all previous actions are provided to the agent directly.

\section{Our Approach}
We list some key process in our algorithm as below:
\begin{itemize}
    \item Train dataset $\mathcal{D}_{train}$ and test dataset ${\mathcal{D}_{test}}$
    \item Apply random augmentation $F$ on image and text
    \item Add the nearest neighbor in support set to augment positive samples
    \item Add active sampling (adversarial or kmeans++) to augment negative samples
    \item Retrieve pair of mined (support set / active sampling) samples as corresponding cross-modal samples
    \item F(Instruction, Observation) to get Action as Follower
    \item H(Observation, Action) to get Instruction as Speaker (different from previous Back translation (s:instructions, t:routes) approach, we reconstruct instruction from observation and action sequences)
    \item Co-training “speaker” and “navigator” via contrastive learning scheme
    \item Compute loss of model
    \item Update model
\end{itemize}

\subsection{Model Overview}
Our proposed architecture is shown in Figure~\ref{fig:vlnMLM}, and the approach can be divided into $4$ steps: 1) Use machine translation augmented data to construct multilingual instructions. 2) Pre-training the Speaker and Navigator module for VLN task. 3) Apply cross-lingual cross-modal contrastive loss in Instruction (by MT-augmentation) and Observation (by spatio/temporal “EnvDrop”). 4) Interactive training of Machine Translation Module and VLN Speaker and Navigator.

\subsection{Encoder}
\textbf{Instruction Encoder}
We follow XLM-R\cite{conneau2020unsupervised} to tokenize an input insturction ${I}$ in language ${l_i}$ to BPE tokens ${t^{l_i} = {t_1^{l_i}, t_2^{l_i}, ..., t_n^{l_i}}}$ using Sentence Piece model. We then project each token to its embedding based on the XLM-R vocabulary and word embeddings. The final representation of each token is obtained via summing up its word embedding, segment embedding, and position embedding as in XLM-R, followed by another Layer Normalization.\\
\textbf{Visual Encoder}
We employ panoramic views as visual inputs to the agent. Each panoramic view consists of ${36}$ images in total. Each image is represented as a ${d-dimensional}$ feature vector. The concatenation of visual feature, output by a Residual Network of the image and a orientation feature vector.\\
\textbf{Action Encoder}
The action encoder is a look-up table, which maps actions types into action embeddings.\\
\textbf{Cross-modal Encoder}
We then apply the multi-layer transformer encoder as a cross-modal encoder on the language embeddings ${h^x_{1:L}}$, visual embeddings ${h^v_{1:T}}$ and action embeddings ${h^a_{1:T}}$. The multimodal encoder returns output embeddings ${(z^x_{1:L}, z^v_{1:T}, z^a_{1:T})}$.
\textbf{Fusion}
We need to fuse vision, language and action representations together to contextualize our instruction-following agent into environmental signals and human intent. Action decision will attend to all language instructions, visual information from panoramic pictures. and history steps. Then scores are transformed into a probability distribution by SoftMax function for a weighted sum of other features.

\subsection{Cross-lingual Navigator}
The output embeddings ${z^v_{1:T}}$ go through a single fully-connected layer to predict agent actions ${\hat{a_{1:T}}}$. During training, the input would be sequence of observation-language paired data and ground truth action sequence. During testing at timestep ${t}$, we input visual observations ${v_{1:t}}$ up to a current timestep and previous actions ${\hat{a_{1:t-1}}}$ taken by the agent. We select the action predicted for the last timestep ${\hat{a_t}}$, ${\hat{c_t}}$ and apply it to the environment which generates the next visual observation ${v_{t+1}}$.

\subsection{Domain-specific Translator}
The translator is based on Multi-lingual Bart. We adopt it for translating from central language to low-resource language. With the interactive training, the translator obtain the knowledge of navigation domain, and provide more accurate translation, which in turn improves the navigation following the instructions of low-resource language.

\subsection{Contrastive Cross-lingual Navigation}
\textbf{Language-level} We construct contrastive pairs in language level. The same instruction with various language are treated as positive samples.
We treat the other $2(N - 1)$ real representation within a minibatch as negative examples. We use cosine similarity to denote the distance between two representation $(I_{l1}, I_{l2})$, that is $\texttt{sim}(I_{l1},I_{l2}) = \mathbf{I_{l1}}^T \cdot \mathbf{I_{l2}}/||\mathbf{I_{l1}}|| \cdot ||\mathbf{I_{l2}}||$. The loss function for a positive pair of examples $(I_{l1}, I_{l2})$ is defined as:
\begin{equation}
\mathcal{L}_{uv} = -\text{log} \frac{\text{exp}(\texttt{sim}(I_{l1}, I_{l2_{i}})/\tau)}{\sum_{\substack{j=1 \\ j\neq i}}^{N} \text{exp}(\texttt{sim}(I_{l1}, \tilde{I_{l2_{j}}})/\tau)}
\label{eq:language_contrastive}
\end{equation}
where $\tau$ denotes a temperature parameter that is empirically chosen as $0.1$.\\
\textbf{Path-level} We construct contrastive pairs in path level, the instructions from the same path are treated as positive samples, and the rest of other instructions in the same mini-batch are treated as negative samples. Similarly, the path-level contrastive loss can be calculated by replacing language $l_1$, $l_2$ to path ${p}$ and augmented path $\hat{p}$.

\section{Preliminary Experiments}
\begin{table}[t]
\resizebox{\textwidth}{!}{%
\centering
\begin{tabular}{l|cccccccc} 
     \text { Split }& \text { Method } & \text { PL } & \text { NE } $\downarrow$ & \text { SR } $\uparrow$ & \text { SPL } $\uparrow$ & \text { SDTW } $\uparrow$ & \text { NDTW } $\uparrow$ \\
    \hline 
    \multirow{3}{*}{Train} & \text {VMLN, M2M} & \text{12.1 } & \text{6.20 } & \text{47.2 } & \text{45.4 } & \text{39.9 } & \text{59.3 } \\
    \text {} &\text {Pivot, M2M} & 13.4 & 10.5 & 27.0 & 25.1 & 22.2 & 40.7 \\
    \text {} & \text {Upper bound} & 15.0 & 4.9 & 60.3 & 57.4 & 51.7 & 65.9 \\
    \hline 
    \multirow{3}{*}{ Val-Seen } & \text {VMLN, M2M} & \textbf{12.9 } & \textbf{11.3 } & \textbf{23.4 } & \textbf{22.0 } & \textbf{20.3 } & \textbf{41.4 } \\
    \text {} &\text {Pivot, M2M} & 13.7 & 11.4 & 22.2 & 20.3 & 19.3 & 37.9 \\
    \text {} & \text {Upper bound} & 15.8 & 8.1 & 40.2 & 36.7 & 33.7 & 52.8 \\
    \hline 
    \multirow{3}{*}{ Val-Unseen } & \text {VMLN, M2M} & \textbf{11.4} & \textbf{10.8} & \textbf{24.4} & \textbf{21.5} & \textbf{19.5} & \textbf{41.0}\\
    \text {} & \text {Pivot, M2M} & 12.6 & 10.8 & 21.5 & 20.0 & 17.7 & 37.9 \\
    \text {} & \text {Upper bound} & 15.2 & 8.0 & 36.3 & 33.1 & 30.6 & 51.2 \\
    \end{tabular}
    }
    \caption{Preliminary results of our Vision-Multilanguage Navigation (VMLN) method and its upper bound on English (En) and Hindi (Hi). Best performance is highlighted in bold.}\label{table:resultM2M}
\end{table}

\begin{table}[t]
\centering
\resizebox{\textwidth}{!}{%
\begin{tabular}{l|cccccccc} 
    \text { Split }  & \text { Method } & \text { PL } & \text { NE } $\downarrow$ & \text { SR } $\uparrow$ & \text { SPL } $\uparrow$ & \text { SDTW } $\uparrow$ & \text { NDTW } $\uparrow$ \\
    \hline 
    \multirow{3}{*}{ Train } & \text {VMLN, Helsinki} & 13.57 & 6.82 & 44.5 & 42.1 & 36.3 & 55.3 \\
     \text {} & \text {Pivot, Helsinki} & 7.9 & 11.3 & 17.8 & 16.8 & 14.2 & 34.6 \\
    \text {} & \text {Upper bound} & 15.0 & 4.9 & 60.3 & 57.4 & 51.7 & 65.9 \\
    \hline 
    \multirow{3}{*}{ Val-Seen } &  \text {VMLN, Helsinki} & 15.25 & 13.26 & \textbf{16.4} & \textbf{15.1} & \textbf{13.4} & \textbf{33.2} \\
    \text {} & \text {Pivot, Helsinki} & 8.1 & \textbf{11.9} & 14.4 & 13.3 & 11.4 & 32.2 \\
    \text {} & \text {Upper bound} & 15.8 & 8.1 & 40.2 & 36.7 & 33.7 & 52.8 \\
    \hline 
    \multirow{3}{*}{ Val-Unseen } & \text {VMLN, Helsinki} & 12.99 & 12.12 & \textbf{17.4} & \textbf{16.0} & \textbf{14.2} & \textbf{34.8} \\
    \text {} & \text {Pivot, Helsinki} & 7.3 & \textbf{11.1} & 15.0 & 14.1 & 11.9 & 33.7 \\
    \text {} & \text {Upper bound} & 15.2 & 8.0 & 36.3 & 33.1 & 30.6 & 51.2 \\
    \end{tabular}
    }
\caption{Preliminary results of our Vision-Multilanguage Navigation (VMLN) method and its upper bound on English (En) and Hindi (Hi). Best performance is highlighted in bold.}\label{table:resultHel}
\end{table}

In this section, we first introduce the multi-lingual VLN benchmark and metrics to test our algorithms. We next show that a naive pivot method fails to achieve a reasonable performance, motivating the design of a specialized algorithm to tackle the challenge. Finally, we show that our proposed novel VMLN framework can achieve relatively better results.

\subsection{Dataset and Metrics}
We evaluate our methods on the newly proposed Vision-and-Language Navigation (VLN) dataset \verb|Room-Across-Room| \cite{ku2020room}, which is multilingual (English, Hindi, and Telugu) and larger (more paths and instructions) than other existing VLN datasets. RxR contain 16522 paths in total, which are split: 11089 train, 232 val-seen (train environments), 1517 val-unseen (val environments), and 2684 test. For each navigation path inside the dataset, there are three different corresponding instructions for each language, i.e., nine instructions per path.

Following the RxR~\cite{ku2020room} benchmark, we evaluate the navigation performance using the standard metrics: Path Length (PL), Navigation Error (NE $\downarrow$), Success Rate (SR $\uparrow$), Success weighted by inverse Path Length (SPL $\uparrow$), Normalized Dynamic Time Warping (NDTW $\uparrow$), and Success weighted by normalized Dynamic Time Warping (SDTW $\uparrow$). We refer interested readers to~\cite{anderson2018evaluation} and~\cite{ilharco2019general} for detailed discussion of these metrics.
\subsection{Pivot Method Fails to Work}
In this subsection, we aim to show that a naive pivot method cannot perform well in the RxR benchmark. Specifically, the pivot method trains the state-of-the-art CLIP-ViL model~\cite{shen2021much} with only human English instructions. During testing, English instructions translated from Hindi are taken as input to guide the navigation. To correctly interpret the efficacy of the pivot method, we compare its performance with one of its upper bounds, i.e., when input instructions are human English, which is equivalent to evaluating the trained VLN model with all English instructions from the RxR dataset. To ensure the translation quality, we applied two different translation models, i.e., M2M~\cite{fan2021beyond} and a model from the Helsinki-NLP group\footnote{\url{https://huggingface.co/Helsinki-NLP}}, to conduct Hindi to English translation.

The preliminary results are shown in \autoref{table:resultM2M} and \autoref{table:resultHel}. None of the pivot methods can approach the performance upper bound, demonstrating the necessity of designing a specialized algorithm to solve the XL-XM VLN tasks. Moreover, the pivot method with different translator models exhibits substantially different performance, demonstrating that the quality of the translation significantly influences the performance.

\begin{table*}[t]
\centering
\caption{Experiment results of our pivot method and its upper bound}\label{table:result-pivot}
\resizebox{\textwidth}{!}{%
\begin{tabular}{lcccccccc} 
   \text { Split }  & \text { Method } & \text { PL } & \text { NE } $\downarrow$ & \text { SR } $\uparrow$ & \text { SPL } $\uparrow$ & \text { SDTW } $\uparrow$ & \text { NDTW } $\uparrow$ \\
   \hline \text { Train } & \text {Pivot, M2M} & 13.4 & 10.5 & 27.0 & 25.1 & 22.2 & 40.7 \\
  \text {} & \text {Pivot, Helsinki} & 7.9 & 11.3 & 17.8 & 16.8 & 14.2 & 34.6 \\
   \text {} & \text {Upper bound} & 15.0 & 4.9 & 60.3 & 57.4 & 51.7 & 65.9 \\
   \hline \text { Val-Seen } & \text {Pivot, M2M} & 13.7 & 11.4 & 22.2 & 20.3 & 19.3 & 37.9 \\
   \text {} & \text {Pivot, Helsinki} & 8.1 & 11.9 & 14.4 & 13.3 & 11.4 & 32.2 \\
   \text {} & \text {Upper bound} & 15.8 & 8.1 & 40.2 & 36.7 & 33.7 & 52.8 \\
   \hline \text { Val-Unseen } & \text {Pivot, M2M} & 12.6 & 10.8 & 21.5 & 20.0 & 17.7 & 37.9 \\
   \text {} & \text {Pivot, Helsinki} & 7.3 & 11.1 & 15.0 & 14.1 & 11.9 & 33.7 \\
   \text {} & \text {Upper bound} & 15.2 & 8.0 & 36.3 & 33.1 & 30.6 & 51.2 \\
   \end{tabular}
   }
\end{table*}

\subsection{Preliminary Results}
Preliminary results are shown in \autoref{table:resultM2M} and \autoref{table:resultHel}. The pivot method failed to successfully model the navigation instructions directly translated from Hindi, on both translation settings(M2M and Helsinki). When the VMLN was applied, we can see an obvious improvements on both settings on almost all-level of matrices. In particular, the translation quality from M2M is much better than Helsinki and we also see more gains on M2M. More efforts are needed to confirm the same conclusions from Telugu. We leave this for future work.

\subsection{Analysis}
Our proposed VMLN achieves consistent improvement over M2M and Pivotal method. The boosting ability can be due to the consistency learned by aligning English and Hindi together in the shared encoder.
However, more solid experiments are required to confirm our findings. For example, one may argue that the performance gain also benefits from more training data rather than aligned encoder. Also, it is still unclear whether our proposed approach works for 'zero-shot' case. We plan to verify this hypothesis by validating on Hindi and Telugu in the future.

\section{Conclusion}
We propose a machine-translation augmented framework \methodname~ (Vision-Multilanguage Navigation) for cross-lingual vision-language navigation.
We first collect multilanguage vision-language-navigation dataset via machine translation.
Then we extend the standard VLN training objectives to multilingual setting via a cross-lingual language encoder.
The alignment between different languages is captured through shared visual and action context via cross-modal transformer, which encodes the inputs of language instruction, visual observation and action decision sequences.
Experimental results over Room Across Room Dataset show superiority of our method and point to a new research direction of real-world multi-lingual navigation.

\bibliographystyle{IEEEbib}
\bibliography{strings,refs}
\end{document}